\documentclass[conference]{IEEEtran}
\IEEEoverridecommandlockouts
\usepackage{gensymb}
\usepackage{cite}
\usepackage{chemformula}
\usepackage{amsmath,amssymb,amsfonts}
\usepackage{algorithmic}
\usepackage{graphicx}
\usepackage{textcomp}
\renewcommand\IEEEkeywordsname{Keywords}
\usepackage{xcolor}
\def\BibTeX{{\rm B\kern-.05em{\sc i\kern-.025em b}\kern-.08em
    T\kern-.1667em\lower.7ex\hbox{E}\kern-.125emX}}
\begin{document}

\title{Analysis of NARXNN for State of Charge Estimation for Li-ion Batteries on various Drive Cycles}

\author{\IEEEauthorblockN{Aniruddh Herle$^{\alpha}$, \textit{Student Member},  Janamejaya Channegowda$^\beta$, \textit{Member, IEEE}, Kali Naraharisetti$^{\gamma}$, \textit{Member, IEEE} \\
$^{\alpha}$Ramaiah Institute of Technology, Bangalore, India, aniruddh.herle@gmail.com \\
$^\beta$Ramaiah Institute of Technology, Bangalore, bcjanmay.edu@gmail.com}
$^\gamma$Infineon Technologies, swaraj.kali@gmail.com}

\maketitle

\begin{abstract}
Electric Vehicles (EVs) are rapidly increasing in popularity as they are environment friendly. Lithium Ion batteries are at the heart of EV technology and contribute to most of the weight and cost of an EV. State of Charge (SOC) is a very important metric which helps to predict the range of an EV. There is a need to accurately estimate available battery capacity in a battery pack such that the available range in a vehicle can be determined. There are various techniques available to estimate SOC. In this paper, a data driven approach is selected and a Non-linear Autoregressive Network with Exogenous Inputs Neural Network (NARXNN) is explored to accurately estimate SOC. NARXNN has been shown to be  superior to conventional Machine Learning techniques available in the literature. The NARXNN model is developed and tested on various EV Drive Cycles like  LA92, US06, UDDS and HWFET to test its performance on real-world scenarios. The model is shown to outperform conventional statistical machine learning methods and achieve a Mean Squared Error (MSE) in the $\mathbf{10^{-5}}$ range.
\end{abstract}


\section{Introduction}

Li-ion batteries have become very popular for use in vehicles due to several desirable traits \cite{Sidhu2019}. Most importantly, they produce clean energy, a growing need of the day. Managing these batteries require sophisticated systems, as any over-charging or over-discharging can severely damage the battery health, leading to catastrophic explosions \cite{How2019}. State of Charge (SOC) estimation is a key part of the operation of the Battery Management System in Electric Vehicles, and must be estimated with the help of strongly correlated quantities, like voltage, current and temperature as it cannot be measured directly \cite{How2019}. A lot of research has shown that machine learning approaches are successful in this task.\\
\indent In \cite{Bermejo2018}, a NARX model was trained on data gathered from a 2016 Nissan Leaf’s 52 trips covering a distance of around 800 km. Their model achieved an accuracy of 1 $\times{10^{-6}}$ when tested on the same car. The same year, NARX was used together with Particle Swarm Optimisation algorithm for the estimation of SOC from voltage, current and temperature at different temperatures in Lipu at al \cite{Lipu2018}. They used the US06 drive cycle for this work, and achieved a minimum value of MSE of 3.8 $\times{10^{-5}}$. They had an optimal value of 5 input delays, and 17 hidden neurons for their final network architecture for 25 \degree Celsius. In 
Chuangxin et al \cite{Chuangxin2019}, a NARX NN was used in conjunction with a Genetic Algorithm, and a model with an input delay of 2, feedback delay of 10 and 6 hidden layers was found to perform best with an MSE of 2.0995 $\times{10^{-5}}$ and MAE of 0.0033. The model was trained and evaluated on 109 discharge cycles of a B0006 battery. Then, in Hannan et al \cite{Hannan2020}, Recurrent nonlinear autoregressive with exogenous inputs (RNARX) was used together with Lightning Search Algorithm to estimate the SOC of a LiNCA battery. They showed that RNARX with LSA outperformed other hyper parameter tuning techniques like BSA, PSO and GSA. NARXNNs have been seen to grow in popularity in the recent years, and in Abbas et al \cite{Abbas2019}, it was even shown that NARXNN outperformed the most popular SOC estimation model, LSTM-RNN, in terms of RMSE. They trained the models using the UDDS Drive cycle.\\
\indent As was evident in the literature survey, a common basis of training and testing these models is not present. In this work, a NARX Neural Network is trained and evaluated on various drive cycles, and its performance across these is analysed. This is important as each drive cycle represents a different real-life scenario the model might face. As can be seen in the research already done, NARX models have not been tested on a diverse range of possible real-life situations.\\
\indent This is an issue which is addressed in this work. Here, several machine learning models are trained on a hybrid SOC variation curve made from several combined drive cycles from the dataset in \cite{Chemali2018}. The dataset was compiled using a Panasonic 18650PF battery, with a Rated capacity of 2.9 Ah, Nominal Voltage of 3.6 V and a Maximum Voltage of 4.2 V. In this work, the data collected at 25 \degree Celsius and a 1 C rate is used. The NARX models with 4 Hidden Neurons with 50 and 100 delay units performed best, and their performance was analysed on various drive cycles. \\

\begin{figure*}[t]
\begin{center}
\includegraphics[scale=0.75]{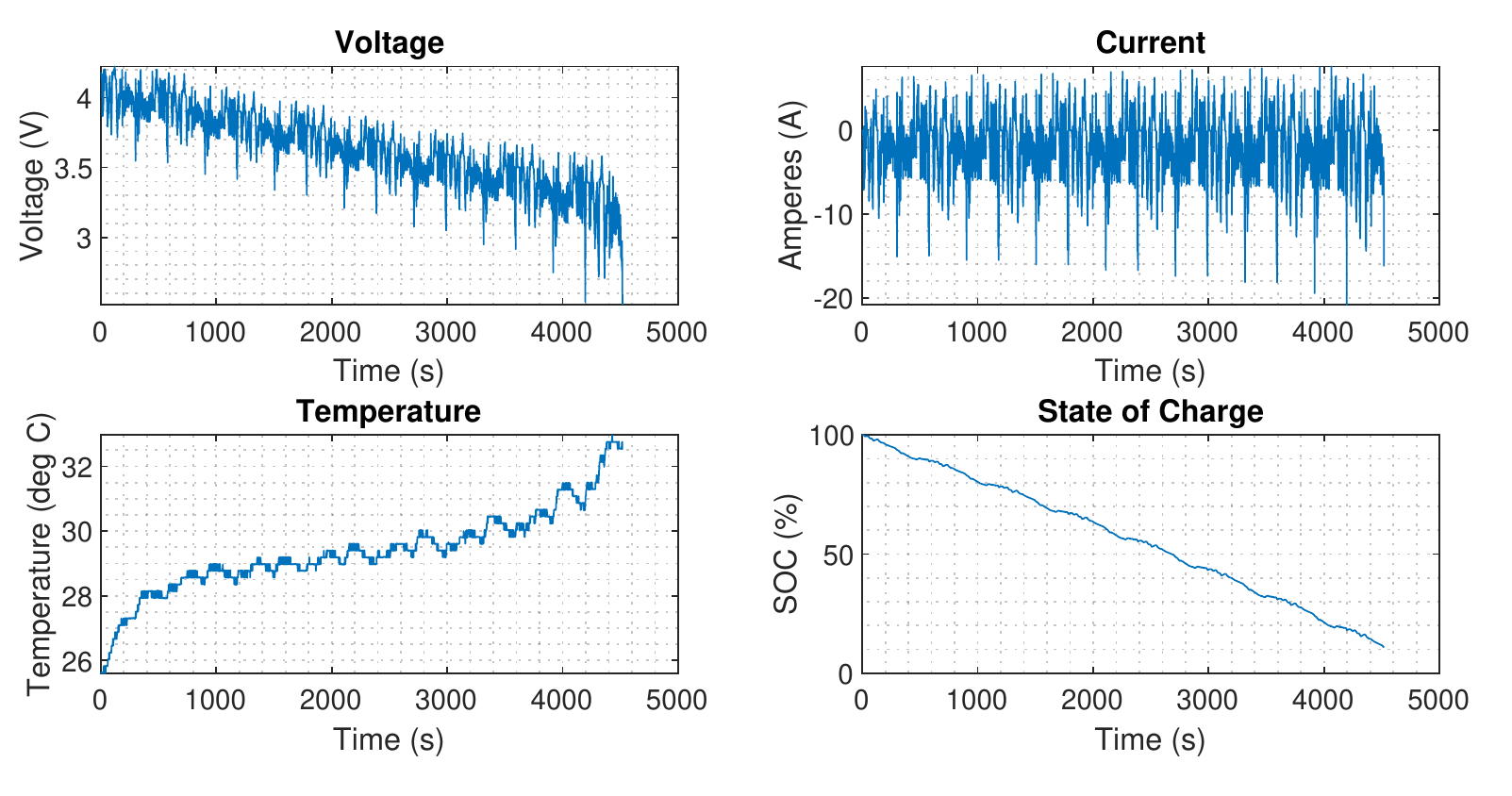}
\vspace*{-0.5cm}
\caption{Variation of Voltage, Current, Temperature and SOC of US06 Drive Cycle at 25 \degree Celsius}
\end{center}
\label{fig:1}
\end{figure*}

\section{Dataset}

\indent The dataset from \cite{Chemali2018} was used for both training and testing the NARXNN used in this work. Digatron Firing Circuits testing equipment was used to collect this data from a Panasonic 18650 battery cell with a lithium nickel cobalt aluminium oxide (\ch{LiNiCoAlO2}) chemistry. The cell was subjected to a variety of drive cycles at different temperatures (0 \degree C to 25 \degree C). The battery specifications are shown in Table I and the variation of cell parameters is shown in Fig 1.\\ 
\indent For this work, the 25 \degree Celsius case is considered. This temperature was chosen as it allows for regenerative braking data to also be collected during testing, as charging is only safe at temperatures greater than 10 \degree C \cite{Chemali2018}. This allows regenerative braking dynamics to also be captured by the neural network. The current sensor used for the data acquisition has an error less than 25 mA, summing up to less than 40 mAh for the entire dataset, which represents only 1.38\% of the battery capacity. In order to test the model on realistic situations, four drive cycles were used, namely HWFET, LA92, US06 and UDDS. 
\begin{table}[hbt!]
\begin{center}
\caption{Panasonic 18650PF Battery Specifications} 
\begin{tabular}{ |c|c| } 
 \hline
 Nominal Capacity & 2.9 Ah \\
 \hline
 Nominal Voltage & 3.6 V \\
\hline 
 Minimum Voltage & 2.5 V \\
\hline 
 Maximum Voltage & 4.2 V \\
\hline
 Power Density & 206.25 Wh/kg\\
 \hline 
\end{tabular}
\vspace*{3mm}

\label{table:1}
\end{center}
\end{table}
\section{NARX}
\indent NARX (Non-linear Autoregressive with Exogenous inputs) Neural Network is a subclass of recurrent neural networks which works well for non-linear time series predictions. It is an improvement on conventional RNNs \cite{Parmezan2019}. It outperforms conventional RNN in training time, accuracy and generalisation to new data \cite{Chuangxin2019}. NARX attempts to model in time the past values of the time series that it is trying to predict, as well as other variables that affect the target (exogenous inputs).\\
\indent The NARXNN attempts to map State-of-Charge in terms of Voltage, Current and Temperature. This is done using an internal neural network \cite{Boussaada2018} as shown in Fig 4. Artificial Neural Networks are especially good at accurately mapping non-linear relationships between the input and output. Each value of the input is multiplied with the weights and summed with the bias value to get the output at each layer. It is these weights and biases that are optimized during the training process. What sets a NARXNN apart from a straightforward ANN is that it also takes into account the past values of the target parameter. This means that there is an in-built memory feature in this architecture \cite{Sun2020}.\\
\indent This memory is provided by the delay units present in the model. By varying the number of delays, the number of time-steps into the past that the model learns can be controlled. In essence, the parameters that control the architecture of the model are the number of delay units and the number of hidden neurons. The delay units represents how many time steps into the past values of Voltage, Current and Temperature and SOC the model takes as input, and the number of hidden neurons represents the number of neurons present in the hidden layer. As can be seen in Fig. 4, 100 delay units means that 100 past values of the 3 inputs ($x(t)$) and the SOC value ($y(t)$) are taken to train the model.\\
\indent Training was done in an open-loop using the Levenberg–Marquardt technique. This technique allows the neural network to converge to a solution, in a fast and stable way. Training could have been done in either an open-loop fashion or a closed-loop fashion. Closed-loop method would be used when the model is deployed in an EV, where the model's predictions are fed-back as inputs to the model for the next prediction i.e for the past values of SOC. As the true SOC values are available for training, these are used as the past values while training the model, which is why it is classified as open-loop training.

\begin{figure*}[htbp]
\centering
\begin{minipage}{.45\textwidth}
\includegraphics[scale=0.4]{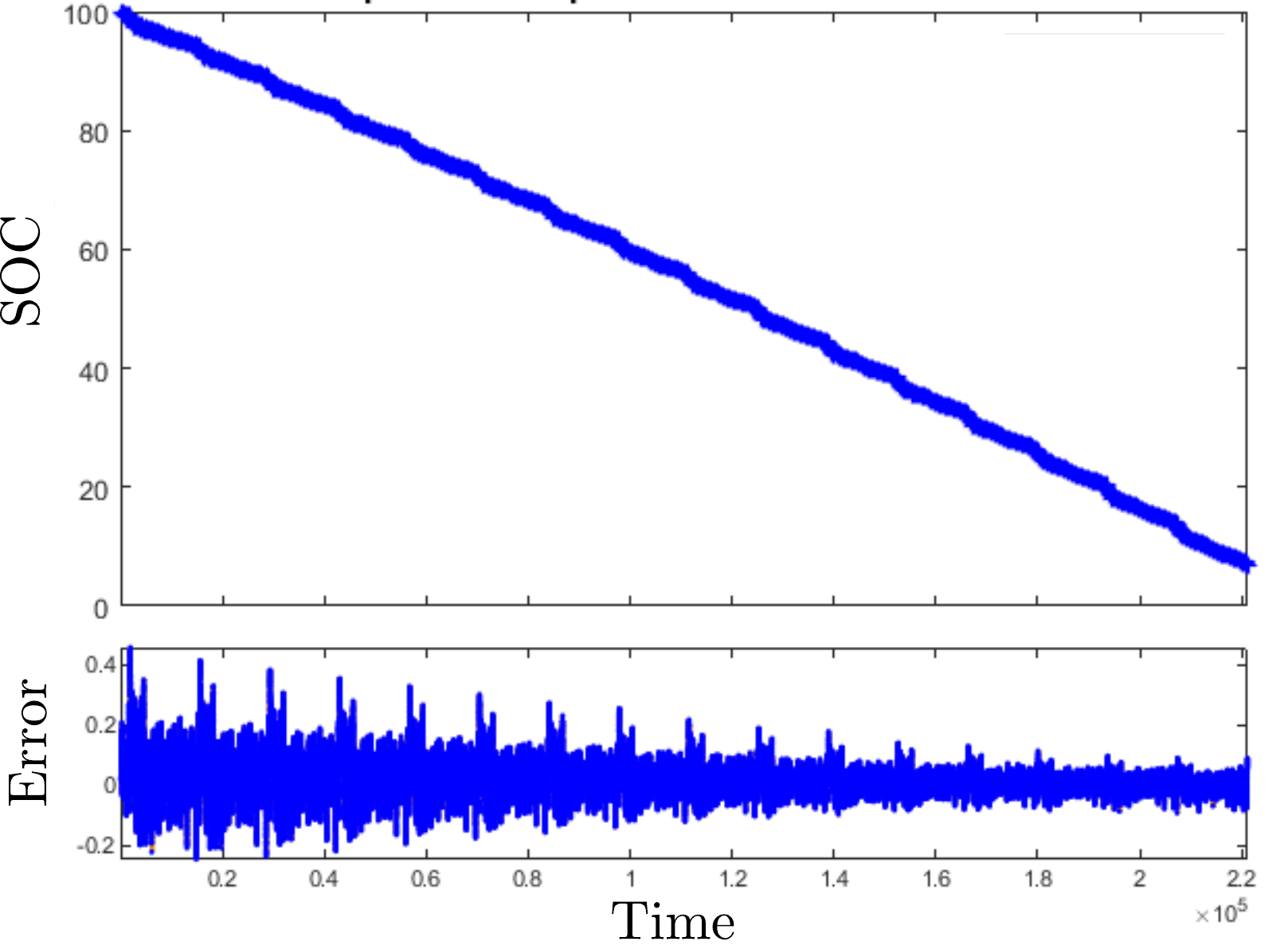}
\caption{4 N, 50 D Test on UDDS}
\label{fig:3}
\end{minipage}\qquad
\begin{minipage}{.45\textwidth}
\includegraphics[scale=0.4]{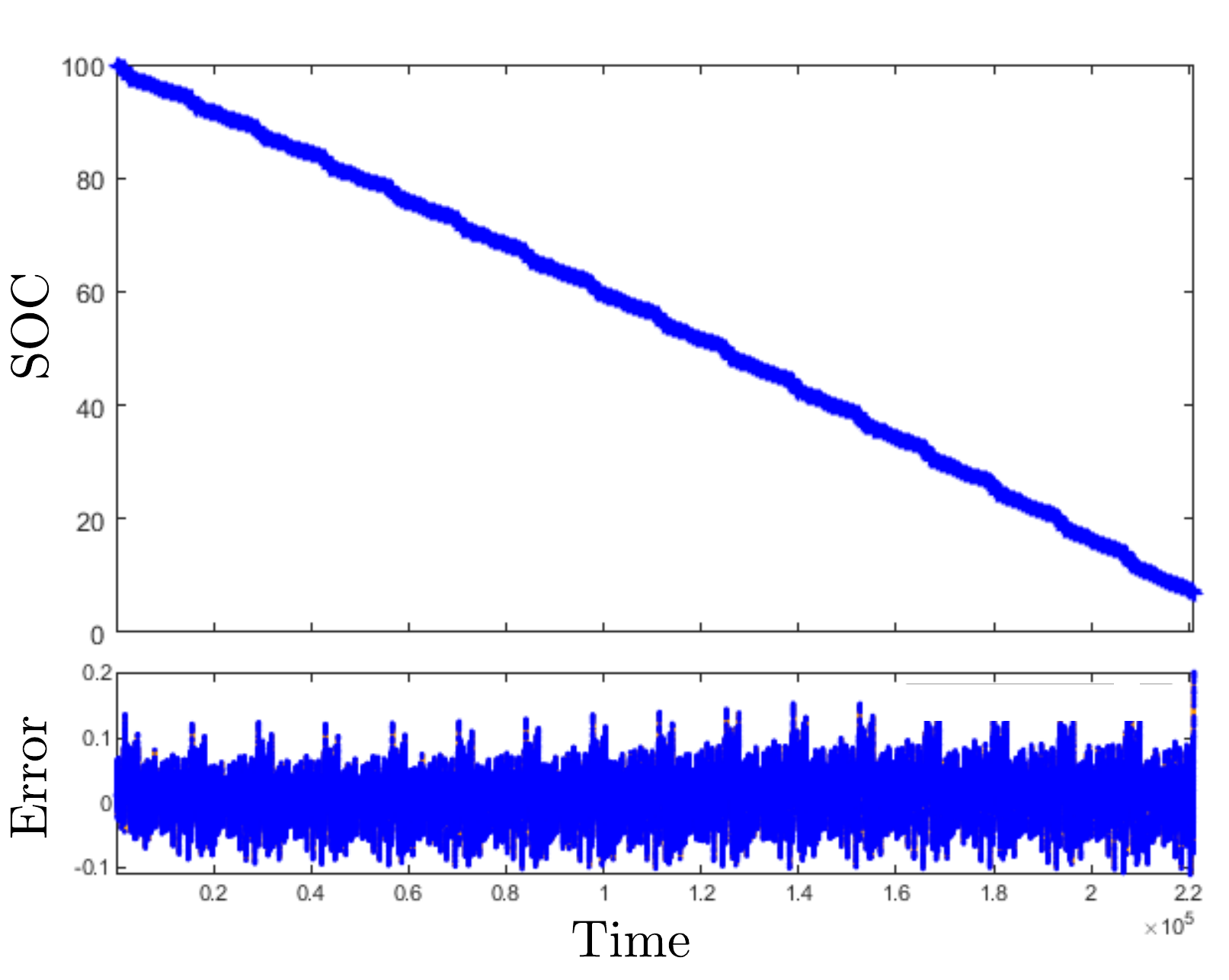}
\caption{4 N, 100 D Test on UDDS}
\label{fig:4}
\end{minipage}\qquad
\end{figure*}

\begin{figure}[hbt!]
\begin{center}
\includegraphics[scale=0.5]{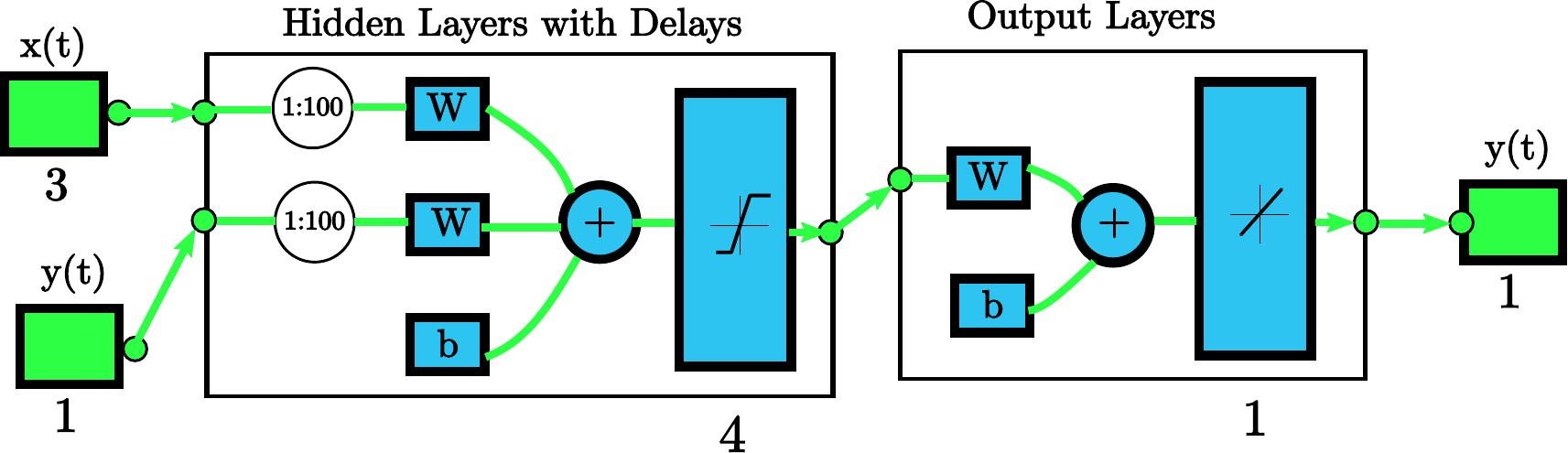}
\caption{NARX Model Architecture with 100 delays and 4 Hidden Neurons}
\end{center}
\label{fig:5}
\end{figure}

\begin{table*}
\parbox{.45\linewidth}{
\centering
\caption{NARX Model Performance}
\begin{tabular}{ccc}
\hline
{Hidden Neurons} & {Delay Units} & {MSE ($\times 10^{-5})$} \\       
     \hline
4 & 2 & 3.31161  \\
  \hline
4 & 20 & 3.12368 \\
\hline
\textbf{4} & \textbf{50} & \textbf{2.99242}\\
\hline
\textbf{4} & \textbf{100} & \textbf{3.03910}\\
\hline
\end{tabular}
\vspace*{3mm}

}
\hfill
\parbox{.45\linewidth}{
\centering
\caption{NARX Model MSE for Drive Cycles}
\begin{tabular}{ccc}
 \hline
{Drive Cycle} & {4 N, 100 D} & {4 N, 50 D}  \\       
  \hline
HWFET & 1.04043$\times 10^{-3}$ & 3.32609$\times 10^{-3}$ \\ 
   \hline
LA92 & 1.07019$\times 10^{-3}$ & 3.36137$\times 10^{-3}$ \\ 
   \hline
UDDS & 1.07284$\times 10^{-3}$ & 3.72901$\times 10^{-3}$ \\ 
   \hline
US06 & 0.947019$\times 10^{-3}$ & 2.72715$\times 10^{-3}$ \\ 
   \hline
\end{tabular}
\vspace*{3mm}

}
\end{table*}

\section{Results}
 MATLAB 2019b was used to train and test the models \cite{MATLAB:2019}. Table II shows the compiled results from many possible combinations of neurons and delays. The NARX model with 4 hidden neurons performed best and the MSE variation with number of delays is shown. In Table \ref{table:2}, the best performing statistical models are shown, trained on a hybrid drive cycle SOC variation curve made from several drive cycles. From all the statistical models tested, Tree (Fine) and Gaussian Process Regression (Matern 5/2, Rational Quadratic, Exponential) performed best in terms of MSE. It can be seen clearly that the NARX method far outstrips the statistical methods, providing strong evidence for the fact that taking temporal variations into account is vital for the SOC estimation task. The model with 4 Hidden Neurons and 50 delays performed best. \\
\indent Previous work \cite{Hussein2014} has shown that increasing the number of hidden neurons does not significantly improve performance. As seen in Table II, an increase in the number of delay units causes a decrease in the MSE. This is because the model takes more time steps into account. 50 delay units means that the model estimates the new value taking the last 50 values of current, voltage, temperature and SOC as inputs.\\
\indent The performance of the 100 delay and 50 delay units model during training was comparable, but when tested on various drive cycles (LA92, HWFET, UDDS and US06), the 100 delay units model outperformed the 50 delay units model for each drive cycle, by a factor of roughly 3. This can be seen in Table III, which shows the MSE values on each of the drive cycles. This shows that taking more time variations into account is beneficial to the predictive power of the model.\\
\indent In Figs. \ref{fig:3} and \ref{fig:4}, the Predicted vs Actual values of the 50 delay and 100 delay units model on the UDDS Drive Cycles, with SOC and Error in percentages, is shown. It shows the near perfect fit between the Actual values, and the predicted value of the respective NARXNN model. The error can be seen to vary with the time instances as well. The error for the 50 delay model is seen to decrease with time, whereas the 100 delay model stays roughly constant. This could mean that the 50 delay could outperform the 100 delay model at lower SOC regions, which is a major concern in SOC estimation. Overall, however, the error for the 100 delay model on the US06 was the least, indicating that this model would perform better than the 50 delay model in US06-like real-world scenarios.\\ 

 \begin{table}[htb!]
  \begin{center}
  \caption{Table of Statistical ML Methods Performance}
    \begin{tabular}{|c|c|c|c| }
 \hline
{\textbf{Model}} & {RMSE} & \textbf{$R^2$} & {\textbf{MSE}}  \\       
  \hline
 Fine Tree & 2.4754 & 0.99 & 6.1278  \\ 
 \hline
 Matern 5/2 Gaussian Process Regression & 2.0922 & 0.99 & 4.3722 \\ 
 \hline
 Rational Quadratic Gaussian Process Regression & 1.8854 & 1.00 & 3.5347 \\ 
 \hline
 \textbf{Exponential} \textbf{Gaussian Process Regression} & \textbf{1.7898} & \textbf{1.00} & \textbf{3.2397}\\ 
 \hline
   \end{tabular}
  \end{center}

\label{table:2}
\end{table}
\section{Conclusion}
In this work, the performance of a NARX model was tested on several different drive cycles, and a Neural Network with 4 layers and 100 delay units was found to perform best overall. The models were trained in an open-loop fashion using the Levenberg-Marquardt technique, and testing was done on various key drive cycles. This was done in order to ascertain whether such a method is feasible for real-life deployment by testing it on data similar to that it will see in an EV.\\
\indent From the trend in the data, it can be inferred that taking more delay units could further improve accuracy. An interesting observation was made regarding the decrease in error in time for the 50 delay model, indicating that it could potentially be more accurate for low prediction in lower SOC regions. Further work can be done to study the effect of further increasing the delay units. This work also gives credence to the idea of using two separate models in each region of the SOC curve, a 100 delay model in the high SOC regions and a 50 delay model in the lower SOC region. This could be a topic for future work.

\bibliographystyle{ieeetran}
\bibliography{references}

\end{document}